\documentclass[lettersize,journal]{IEEEtran}
\usepackage{amsmath,amsfonts}
\usepackage{algorithmic}
\usepackage{algorithm}
\usepackage{array}
\usepackage[caption=false,font=normalsize,labelfont=sf,textfont=sf]{subfig}
\usepackage{textcomp}
\usepackage{stfloats}
\usepackage{url}
\usepackage{verbatim}
\usepackage{graphicx}
\usepackage{cite}
\usepackage{url}

\usepackage{breakurl}

\begin{document}

\author{Cliff B.~Abbott, 
        Mark~Elo, 
        and Dmytro A.~Bozhko%
\thanks{Cliff B. Abbott and Dmytro A. Bozhko are with the Department of Physics and Energy Science, University of Colorado Colorado Springs, Colorado Springs, CO 80918, USA}%
\thanks{Mark Elo is with Tabor Electronics, Inc., Hatasia 9, Nesher 3660301, Israel}%
}

\title{Perturbative Gradient Training: A novel training paradigm for bridging the gap between deep neural networks and physical reservoir computing}

\maketitle

\begin{abstract}
We introduce Perturbative Gradient Training (PGT), a novel training paradigm that overcomes a critical limitation of physical reservoir computing: the inability to perform backpropagation due to the black-box nature of physical reservoirs. Drawing inspiration from perturbation theory in physics, PGT uses random perturbations in the network’s parameter space to approximate gradient updates using only forward passes. We demonstrate the feasibility of this approach on both simulated neural network architectures, including a dense network and a transformer model with a reservoir layer, and on experimental hardware using a magnonic auto-oscillation ring as the physical reservoir. Our results show that PGT can achieve performance comparable to that of standard backpropagation methods in cases where backpropagation is impractical or impossible. PGT represents a promising step toward integrating physical reservoirs into deeper neural network architectures and achieving significant energy efficiency gains in AI training.
\end{abstract}


\section{\label{sec:level1}Introduction}
Machine Learning has seen explosive growth in recent years.  With that growth comes an immense increase in the power consumed when training and operating AI/ML models.  Training the ML model Meena had a carbon footprint roughly equal to driving 242,231 miles in an average passenger vehicle \cite{wu2022sustainable} and ChatGPT-4 cost more than \$100 million to train \cite{wiredOpenAIsSays} with the majority of the cost assumed to be energy usage.  Some estimates even put the annual deployment energy consumption for popular models at 25x the energy cost of training \cite{energy1,energy2}, which would put an energy price tag on ChatGPT-4 of around \$2.5 billion.  On the lower end, estimates still indicate that the energy cost of training makes up only 10-40\% of the lifetime cost\cite{george2023environmental,wu2022sustainable}.  The growth in the energy demands for AI has been exponential \cite{george2023environmental} and with current demands for a single model like ChatGPT being equivalent to that of 33,000 households in the US \cite{washingtonQAResearcher}, more energy efficient forms of machine learning will be necessary to continue development in the field. 

Reservoir computing (RC) is a promising concept for reducing the energy costs of machine learning. An RC takes a lower-dimensional input and maps it to a higher-dimensional output via a complex, nonlinear, but dynamically consistent process.  Unlike other machine learning architectures, the training process is regulated to the output or readout layer(s) of the RC, allowing significant reductions in training time \cite{lukovsevivcius2009reservoir}.  The idea of RCs has been previously applied to the transformer architecture, which currently dominates the field of Large Language Models, and showed that the introduction of "frozen layers" as reservoirs had a significant improvement in performance and training time \cite{shen2021reservoir}.  Physical reservoir computing takes this concept and uses other natural systems, such as water waves \cite{waterwaves}, to perform the non-linear process.  Magnonics systems are one such promising example of a physical reservoir computer with significantly lower energy costs, with potential energy savings of up to 90\% compared to traditional electronic systems \cite{Intro1, Intro2, Intro3, Intro4, Intro5, Intro6, Intro7, Intro8, Intro9,Abbott2024,Namiki2024}. The ability to replace conventional electronic systems with magnonic reservoirs could potentially save \$100's of millions of dollars in lifetime energy costs.

Despite their promise, physical RCs face a critical limitation: the reservoir acts as a black box, making internal parameter access and backpropagation impossible.  This severely limits the applicability of physical RCs to only the input layer of a network.  As cutting-edge models are getting deeper and more complex, this means the impact of a physical RC to reduce energy costs is diminishing.  To address this, we propose Perturbative Gradient Training (PGT), a novel method that eliminates the need for backpropagation, enabling physical reservoir computers to be seamlessly integrated into neural networks, regardless of depth and location of the RC.

This paper is structured as follows: Section II introduces the theoretical framework and applications of PGT. Section III presents results, comparing PGT to standard backpropagation in a simulated reservoir system. Section IV applies PGT to an actual physical reservoir system and examines its performance in experimental settings. Finally, Section V discusses the results and highlights the challenges and opportunities for further development of Perturbative Gradient Training.

\section{Theory}
The current state-of-the-art training methods, while differing in how they update model parameters, all rely on backpropagation.  This is where the model is run in a forward pass on a set of training data and a loss is calculated as a comparison of the model output to the target data.
\begin{equation}
 L=f(y,y';\theta)
\end{equation}

Here $f()$ is the chosen function, such as a mean squared difference, $y$ and $y'$ are the model output and target respectively, and $\theta$ is the set of model parameters.  The loss is then passed backwards through the model, layer by layer, calculating the gradient of the loss for each individual parameter in the network and then applying the chosen update function (standard Stochastic Gradient Descent given here):
\begin{equation}
    \theta_{Grad} = \frac{\partial L}{\partial \theta}
\end{equation}
\begin{equation}
    \theta_n = \theta_{n-1} - l_{r} \cdot \theta_{Grad}
\end{equation}
Where $l_{r}$ is the learning rate, a hyperparameter set by the user.  Visually, this method searches the n-D parameter space, looking along a single dimension at a time, and determines whether or not the loss improves with an advance in that direction.  This is already a significant approximation for the ideal path to a minimum\cite{sutskever2013importance, zhang2017rethinking}. For instance, in a three-dimensional landscape, the local gradient might indicate that moving north and east individually reduces the loss, while a combined north-east direction could lead uphill.  The update to the model would then have a negative impact.  However, the likelihood that such a negative update would happen repeatedly is low and this method offers a favorable trade in simplicity over precision.  This simplicity allows computers to perform these steps thousands of times, which has historically led to good results.

The concept for Perturbative Gradient Training comes from a technique often used in physics called Perturbation Theory.  Perturbation Theory approximates solutions to complex problems by starting with a simpler, exactly solvable version of the problem and iteratively introducing small corrections. This approach allows for incremental refinement, enabling solutions to problems that would otherwise be too difficult to solve directly.  Here we start with our simple solution as the initial pre-generated parameters.  A small change to some of the parameters is made and we analyze how that change effected the loss.  If the change was favorable, we keep a portion of the change determined by the optimization step.  If the change was unfavorable, we subtract a portion of that change from the original parameters during the optimization step.

\begin{figure}
    \centering
    \includegraphics[width=0.8\linewidth]{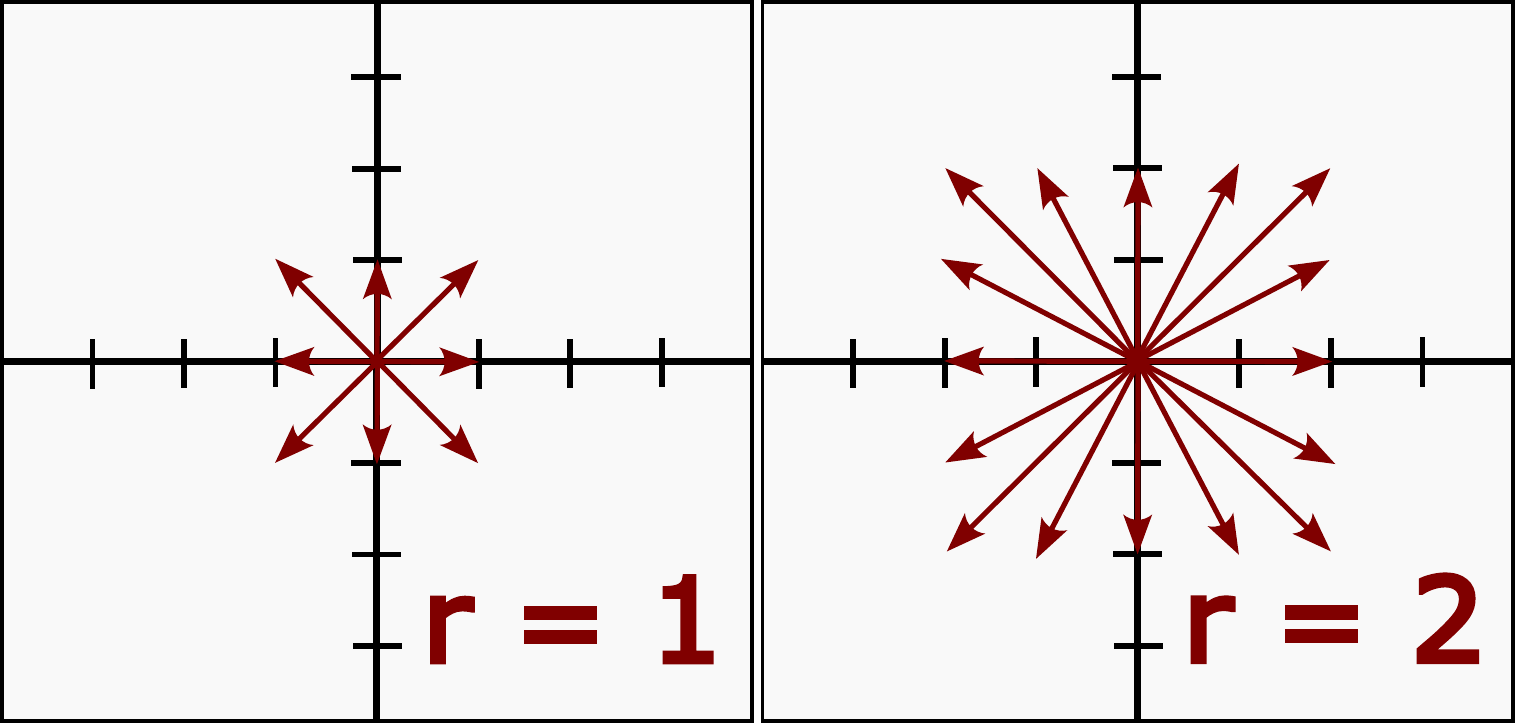}
    \caption{In a 2-D space, $r=1$ corresponds to 8 directions and $r=2$ corresponds to 16 possible directions, demonstrating how changing the range effects the possible directions that PGT can search during training.}
    \label{fig:range}
\end{figure}

In comparison to traditional backpropagation, which looks at each dimension individually in parameter space, PGT picks a single random direction in the overall space and evaluates the gradient of the loss function along that random direction.  This is done by taking loss measurements by a full forward pass of the model just as with equation (1).  However, this time, instead of calculating individual gradients for each dimension, we generate a random perturbation matrix which is a set of integers $[-r, -r+1, ..., 0, ..., r-1, r]$ that is the same size as the parameter space.  This matrix represents the direction in parameter space that we are looking.  Here r is the "range" of the perturbations and is a hyper-parameter set by the user.  Increasing the range increases the number of possible directions that PGT can look in parameter space (ex. Fig \ref{fig:range}).

It is not always beneficial to include all the parameters all the time.  Thus at this step, we introduce the next hyperparameter, the dropout scale.  The dropout scale is a probability factor used to determine whether a specific parameter perturbation is kept or set to zero.  When dropout is set to 0, all perturbations are kept.  At 1, all perturbations are removed.  

Next, we add (subtract) the perturbations to the parameters and determine the new loss.
\begin{equation}
    \theta_{p+} = \theta  + [PM] \cdot \delta
    \end{equation}
\begin{equation}
    \theta_{p-} = \theta  - [PM] \cdot \delta
\end{equation}
Where [PM] is the perturbation matrix and $\delta$ is the scaling factor.  A new loss is then calculated and an overall gradient for that perturbation (i.e. direction in parameter space) is calculated.
\begin{equation}
    L_{p+}=f(y,y';\theta_{p+})
\end{equation}
\begin{equation}
    L_{p-}=f(y,y';\theta_{p-})
\end{equation}
\begin{equation}
    Grad = \frac{L_{p+} - L_{p-}}{2\cdot\delta}
\end{equation}
A "gradient matrix" is then formed by multiplying the gradient by the perturbation matrix.
\begin{equation}
    [Update]=\frac{Grad \cdot [PM]}{[Counts]}
\end{equation}
where [Counts] is a matrix keeping track of the absolute value of the perturbations for scaling.  The update can then be applied using any of the standard optimization step techniques.  This is similar to proposed methods for training Quantum Neural Networks through parameter shifting\cite{ParamShift}.  However, unlike PGT, parameter shifting requires a baseline forward pass and an additional forward pass for every parameter in the network, resulting in potentially millions of forward passes per training sample. With PGT, there are only two forward passes per sample.  During the training cycle, we found that generating a new perturbation matrix for every sample resulted in the best performance.

In this fashion, backpropagation has been replaced by a second forward pass.  Because this method only requires two forward passes, a physical reservoir deployed at any position within the network will not prevent training of the parameters before it.

\section{Simulated Results}
\begin{figure}
    \centering
    \includegraphics[width=0.6\linewidth]{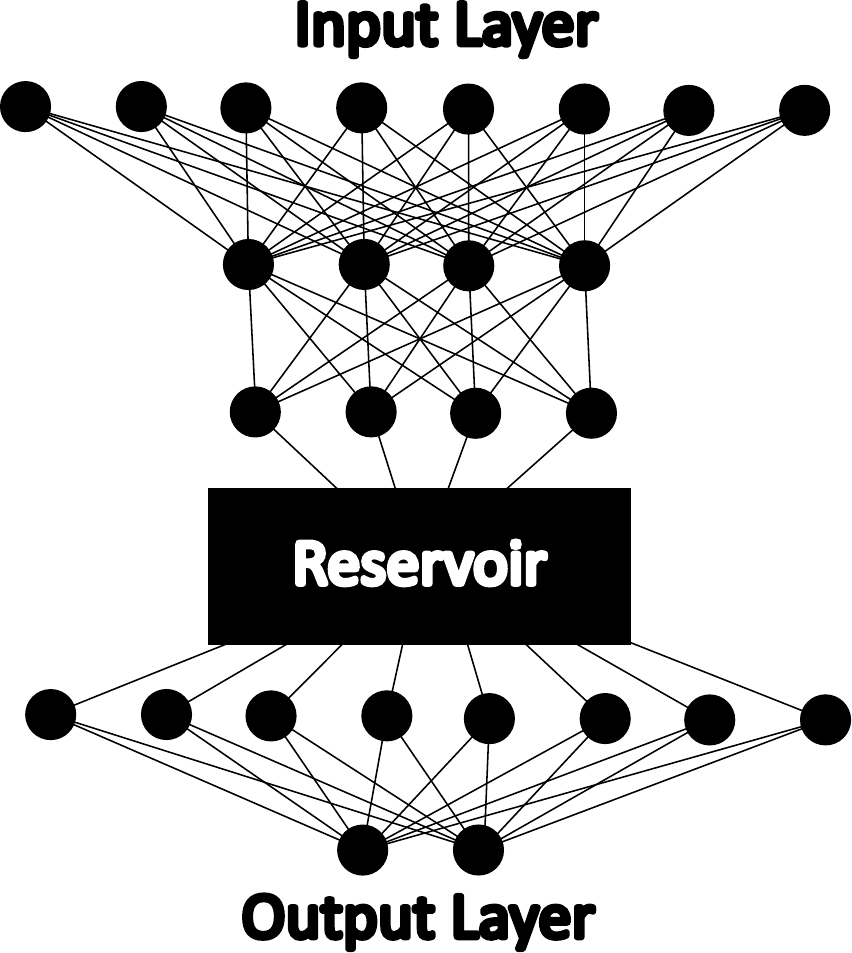}
    \caption{Initial simple network design.  Actual sizes: Input 30, 1st Hidden 200, 2nd Hidden 200, Reservoir In 5, Reservoir Out 100, Output 2.}
    \label{fig:simpleNN}
\end{figure}

For initial testing, we created a small dense network that fed into a simulated physical reservoir that was a second complex network whose parameters were not updated (Fig \ref{fig:simpleNN}) but gradients could be passed back through the reservoir.  PGT is not intended to be used in a system where backpropagation is possible.  However, this comparison is helpful in understanding the trade-off between the benefits of introducing a physical reservoir into the network and the current limitations of this training method.  The simulated reservoir consisted of 4 layers of size 100 or 200.  The forward pass of the reservoir passed through the layers multiple times creating recurrent loops.
\begin{figure}
    \centering
    \includegraphics[width=0.9\linewidth]{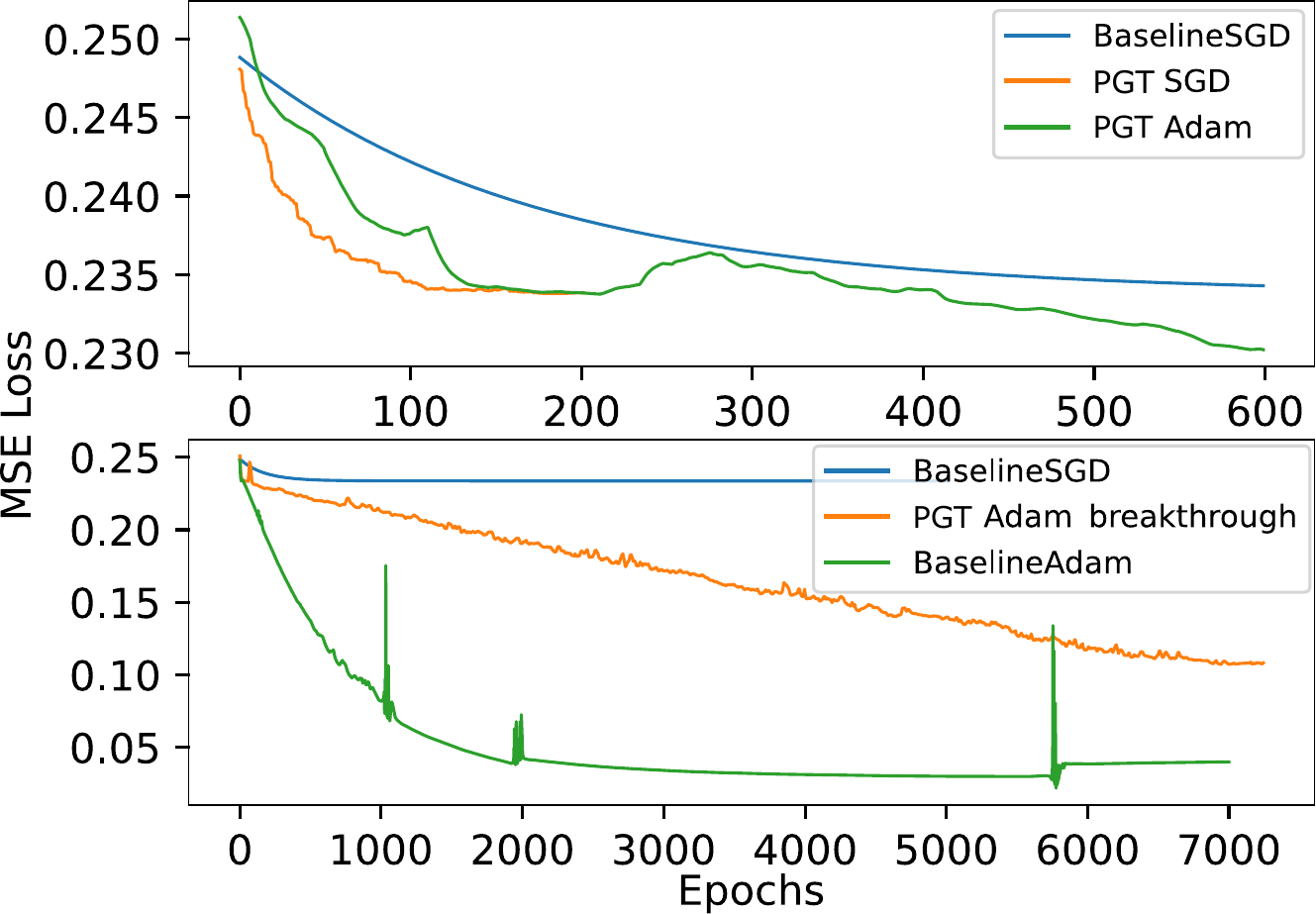}
    \caption{Initial PGT training performance with Stochastic Gradient Descent (SGD) or Adam Optimization compared to backpropagation "Baseline" training.  Top: PGT on the small network outperformed standard SGD significantly both in speed and final loss.  Bottom: Adam backpropagation significantly outperformed SGD and PGT.  Occasionally PGT training with Adam updates would break through the SGD limit and steadily improve towards the best loss achieved by Adam backpropagation.}
    \label{fig:smallNN performance}
\end{figure}
For this training, we used the Wisconsin Breast Cancer Database \cite{breast_cancer_wisconsin_(diagnostic)_17}, which consists of 30 input features resulting in 2 classifications, malignant or benign.  A simple Mean Squared Error Loss was used.
\begin{equation}
   MSE Loss = \sum_n (y_n - y'_n)^2
\end{equation}
This dataset and network design resulted in an interesting outcome in that Stochastic Gradient Descent\cite{SGD} (SGD) backpropagation always stopped at a MSE Loss around 0.2338 whereas Adam\cite{Adam} backpropagation was regularly able to achieve a minimum loss of 0.0221.  When PGT was applied to the problem, both PGT with SGD and Adam optimization achieved results faster than the SGD baseline, but also got stuck at the SGD limit (Fig \ref{fig:smallNN performance} top).  Occasionally, the PGT with Adam optimization would break through that limit and continue steadily towards the min loss achieved by Adam backpropagation.  However, that improvement was very slow comparatively (Fig \ref{fig:smallNN performance} bottom).  Using a low dropout rate, we were able to reach the SGD limit more quickly, but only a high dropout rate (0.999) tended to break through the SGD limit.

\begin{figure}
    \centering
    \includegraphics[width=0.9\linewidth]{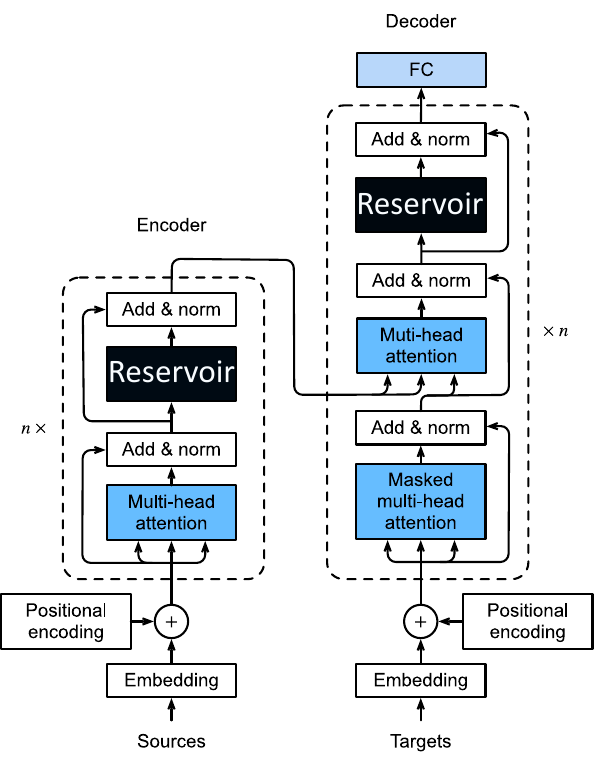}
    \caption{Transformer Architecture with Feed Forward networks replaced by Reservoir Computers.}
    \label{fig:TransArc}
\end{figure}

Next, we looked at applying PGT to the transformer architecture as suggested by Shen et al. \cite{shen2021reservoir} where the feed-forward layers of the transformer are treated as reservoirs (Fig \ref{fig:TransArc}).
We used a single-layer encoder and decoder with an embedding size of 512.  The same simulated reservoir was used as in the previous section.  For this training, we used a small segment of the Multi30k English to German translation database\cite{Multi30k} (250 samples to train and 250 samples to test).  The outputs were left as embeddings and a MSE Loss was used again.
\begin{figure}
    \centering
    \includegraphics[width=0.9\linewidth]{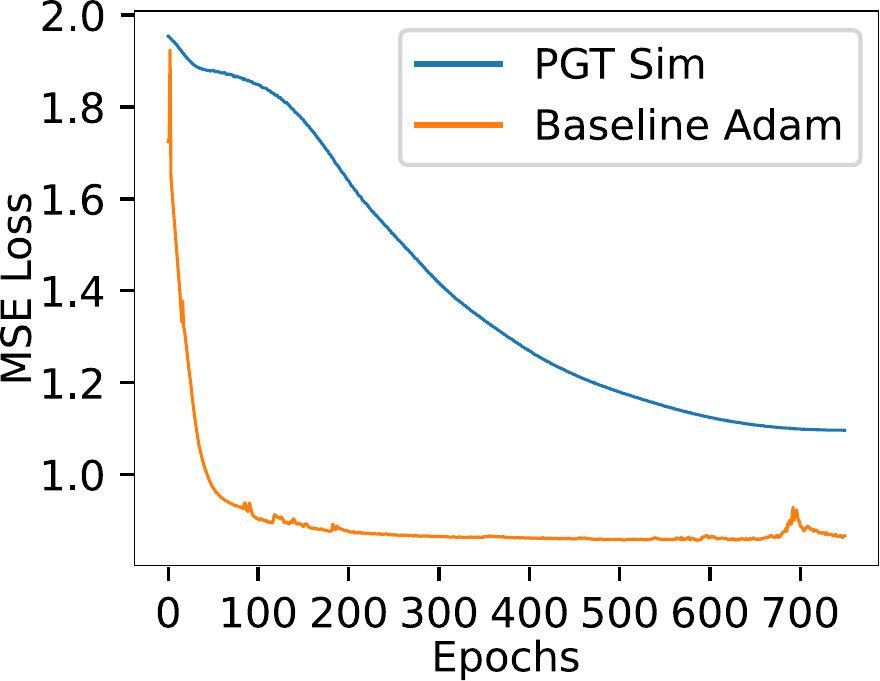}
    \caption{Performance of PGT on a simulated reservoir transformer.  While backpropagation improved significantly faster, PGT was able to get a reasonable result taking only 2.86x longer.}
    \label{fig:TransSim}
\end{figure}
With the transformer, both SGD and Adam backpropagation regularly outperformed PGT.  Only a very high dropout rate of 0.9999, which corresponded to about 330 parameters per perturbation out of 3,363,652 total parameters, was able to consistently train for PGT.  Still, PGT achieved a minimum loss of 1.097 after 745 epochs, taking 2.86x longer to reach maximum performance, while Adam achieved a minimum loss of 0.86 after about 260 epochs (Fig \ref{fig:TransSim}).

\section{Experimental Results}
\subsection{Physical Reservoir Dynamics}

\begin{figure}
    \centering
    \includegraphics[width=0.9\linewidth]{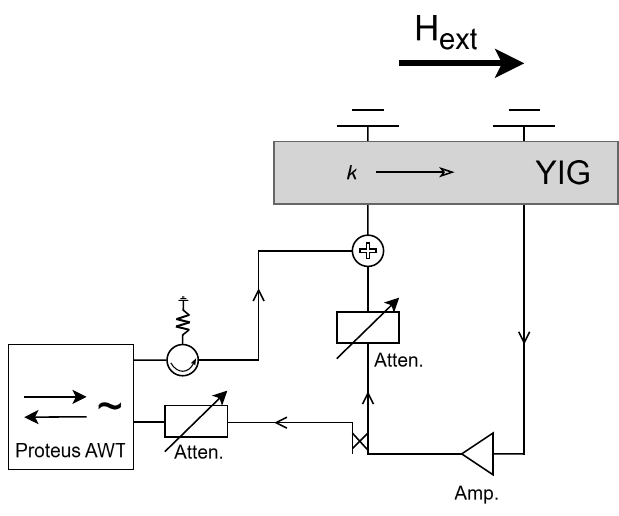}
    \caption{Basic design of the Magnonic Auto Oscillation Ring.  A signal is generated at the AWT Out Channel that travels to the input antenna (gold) of the YIG waveguide.  A spin wave is excited in the waveguide and travels to the output antenna.  The signal generated at the output antenna travels to an amplifier that amplifies the signal by a set amount.  The signal is then split to the In Channel of the AWT and back to the waveguide input antenna via a variable signal attenuator.}
    \label{fig:AORdesign}
\end{figure}

To test our approach experimentally, we used a Magnonic Auto-Oscillation Ring as the physical reservoir computer.  A basic diagram of the setup is provided in Fig.~\ref{fig:AORdesign}.  A thin $10.5\,\mathrm{\mu m}$-thick film of Yttrium Iron Garnet (YIG, Y$_3$Fe$_5$O$_{12}$) is placed into a 500 G external magnetic field applied along the direction of intended wave propagation realizing the so-called Backward Volume Magnetostatic Spin Wave (BVMSW) geometry.  

The orientation of the external magnetic field allows for the excitation of different spin wave geometries and is one of the characteristics that makes magnonics interesting for reservoir computing.  A field applied out of plane of the waveguide excites Forward Volume Magnetostatic Spin Waves (FVMSW) and a field in plane but perpendicular excites a Surface Spin Wave or Damon-Eshbach mode (DE). Each wave type has characteristics that may be of more interest depending on the application.  For example, the propagation direction:  DEs propagate in one principal axis (perpendicular to the in-plane field), but in a nonreciprocal manner (the waves localize differently on opposite surfaces for $+k$ vs. $-k$).  BVMSWs propagate along the axis parallel to the magnetization (i.e., one in-plane axis) in both the $+k$ and $-k$ directions.  FVMSWs propagate in any direction along the 2D plane of the waveguide. Combinations of these waves can be achieved by orienting the external field in between these primary directions.  

A Tabor Proteus Arbitrary Waveform Transceiver (AWT) delivers an electric signal to a thin strip of wire laid perpendicularly across the YIG waveguide, serving as the input antenna. According to Ampère’s Law,
\begin{equation}
    \oint B \cdot dl = \mu_OI_{enc}
\end{equation}
the current in this antenna induces an oscillating magnetic field.  Here B is the Magnetic field along the closed loop path defined by dl, $I_{enc}$ is the current in the antenna, and $\mu_O$ is the permeability of free space.
This oscillating magnetic field excites a spin wave in the magnetic spins of the YIG film.  The dynamics of this wave is governed by the Landau Lifshitz Gilbert (LLG) equation.
\begin{equation}
\frac{d\mathbf{M}}{dt} = -\gamma \mathbf{M} \times \mathbf{H}_{\text{eff}} + \frac{\alpha}{M_s} \mathbf{M} \times \frac{d\mathbf{M}}{dt}
\end{equation}
Here M is the magnetization vector (magnetic moment per unit volume) of the material;  $\mathbf{H}_{\text{eff}}$ is the effective magnetic field, which includes external fields, anisotropy fields, exchange fields, and demagnetization fields; $\gamma$ is the gyromagnetic ratio, which relates the magnetic moment to its angular momentum; $\alpha$ is the Gilbert damping parameter, a dimensionless constant that governs how quickly the magnetization relaxes to equilibrium; and $M_s$ is the saturation magnetization, the maximum magnetization the material can achieve. As the wave travels along the waveguide past the output antenna, the wave creates a current in it in the reverse process as the input antenna.  That signal then travels to an amplifier that increases the signal by a constant amount.  The signal is split back to the AWT for readout and back to the input signal of the input antenna.  In this fashion, a pulse from the AWT will travel around the loop several times (determined by the amount of gain given by the amplifier and attenuator combination on the main ring loop).  Interactions between the looping waves and newly generated waves introduce non-linear dynamics.  This setup gets it's name from the fact that, if the amplification is sufficient, noise in the system will begin to propagate as a signal at the resonant frequency for the waveguide.  The amplification gain at which this occurs is called the "auto-oscillation threshold".  For the purposes of acting as a RC, the ring is operated just below this threshold.  This is done so that previous inputs to the system circulate the loop several times in a decaying fashion, which gives the system it's fading memory, an important characteristic for RCs.  The amount of effective memory can be tuned in the system by changing the amount of gain from the amplifier, which determines how many times a pulse in the system will loop before it decays completely.

The dynamics of the wave propagation, governed by the LLG equation, is the source of the non-linear dynamics required for the Auto-Oscillation Ring to function as the RC.  $\mathbf{H}_{\text{eff}}$ is dependent on the wave amplitude leading to a nonlinear frequency shift and other nonlinear spin-wave phenomena providing a rich set of complex spin-wave interactions.  For more detail into the physics of magnonic auto-oscillation rings and the state of that research, the reader is referred to the reference works by Watt et al.\cite{watt2024numerical,watt2021implementing} and Ustinov et al.\cite{ustinov2024current}.

\subsection{Reservoir Characterization}
A measure of short-term memory (STM) and parity check (PC) are typically used to characterize the ability of auto-oscillation rings to act as physical reservoir computers\cite{STM,PC}. STM is done by giving the RC a series of 1's and 0's in random ordering.  The output of a trained RC is then analyzed to see how many inputs the system can determine were 1 or 0 looking back from the last input to the previous.  For example, a STM of 7 would mean that the RC could reliably tell you the last 7 inputs into the system.  This metric quantifies the "memory depth" of the system.  PC is very similar, but looks at the modulus 2 (XOR) summation of the previous inputs.  This means the RC must keep track of whether the number of 1's in the series of inputs was even or odd.  A high PC score indicates the system's capacity to perform nonlinear transformations.  Typically, scores for STM and PC are given as a Capacity defined as
\begin{equation}
C = \sum_{T_{\text{delay}}=1}^{T_{\text{delay, max}}} \text{Cor}(T_{\text{delay}})^2
\end{equation}
where
\begin{equation}
\text{Cor}(T_{\text{delay}})^2 = \frac{\left(\text{Cov}[y_{\text{train}}(T, T_{\text{delay}}), y_{\text{out}}(T)]\right)^2}{\text{Var}[y_{\text{train}}(T, T_{\text{delay}})] \cdot \text{Var}[y_{\text{out}}(T)]}
\end{equation}
and $T_{delay}$ is how far back you are looking, COV is the covariance, and Var is the variance. This can range up to 10 and 4 for $C_{STM}$ and $C_{PC}$ respectively in numerical simulations\cite{watt2024numerical}, although there is usually a trade off between maximizing one or the other.  The setup for our auto-oscillation ring achieved $C_{STM}=2.91$ and $C_{PC}=0.01$.  For reference, Ustinov et al.\cite{ustinov2024current} had $C_{STM}=5.1$ and $C_{PC}=0.1$ for one of their trials.  Therefore, while it is not state-of-the-art for the field or indicative of the potential for Magnonics Reservoir Computing, our setup was sufficient to conceptually test PGT.
 
\subsection{PGT Results}
\begin{figure}
    \centering
    \includegraphics[width=0.9\linewidth]{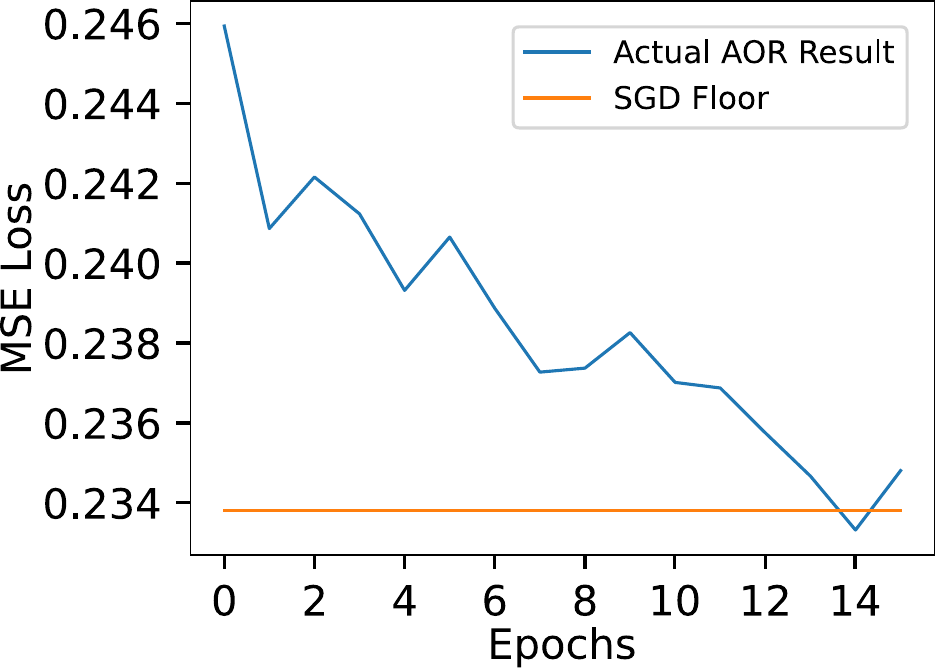}
    \caption{Results for training the initial neural network design with a physical auto-oscillation ring as the reservoir.}
    \label{fig:AORresultSmall}
\end{figure}
Initially we started with the small neural network described by Fig \ref{fig:simpleNN} and replaced the simulated reservoir with the magnonic auto-oscillation ring described above.  This performed extremely well achieving the MSE Loss minimum of SGD backpropagation in only 14 training epochs (Fig \ref{fig:AORresultSmall}).  However, it was noticed during debugging that there was no change in the performance when training only the portion of the network before the reservoir.  Therefore, all the training was occurring in the readout layer post reservoir.  Although still technically successful in the ability of PGT to train the system, it did not prove the ability of using reservoirs deeper in neural networks and training the network in the pre-reservoir portion in a meaningful way.  This is why we shifted to the transformer approach. 

As with the initial neural network, we took the reservoir transformer architecture in Fig \ref{fig:TransArc} and replaced the simulated reservoir with our auto-oscillation ring.  Due to the design and implementation of the ring, each token for each sample had to be sent through the reservoir one at a time which significantly slowed down the training speed of the system.  Even with only 250 training samples and 250 testing samples, out of the 30,000 available, it took around 24 minutes per epoch.  It should be noted that this is a limitation in our realization of the RC with the given hardware and is not a limitation inherent to physical reservoirs in general, nor a limit of PGT.

\begin{figure}
    \centering
    \includegraphics[width=0.9\linewidth]{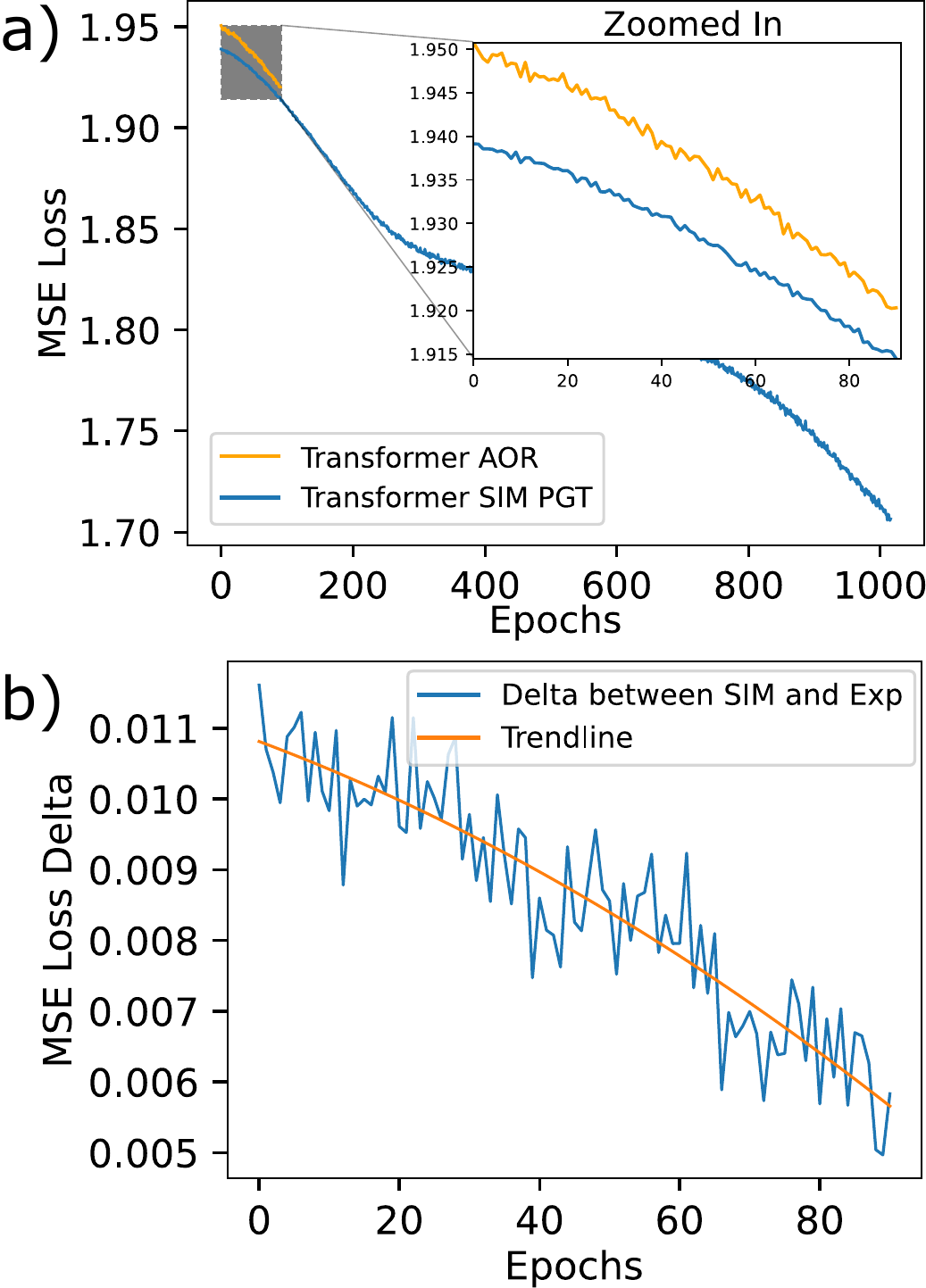}
    \caption{Performance of the physical implementation of the reservoir transformer vs the simulation. a) shows that the physical reservoir is training in a similar fashion to the simulation.  b) the simulation started at a lower loss but the delta is closing showing that the physical implementation is converging faster than the simulation.}
    \label{fig:TransAORcompare}
\end{figure}

The physical reservoir transformer performed as expected from the simulated results (Fig \ref{fig:TransAORcompare}).  Compared to the simulation, we see that the initial loss was higher (simply a result of the parameter initialization) but improves at a consistent rate. Fig. \ref{fig:TransAORcompare}b shows that difference between the simulated and experimental loss is also decreasing with a negative second derivative.  This indicates that the experimental loss appears to be converging slightly faster than simulation, which was also observed in the small initial neural network.

\section{Discussion}
In this work, we introduced the framework for Perturbative Gradient Training (PGT) and demonstrated its potential to address a major limitation in physical reservoir computing. By enabling training without requiring backpropagation, PGT offers a viable solution for systems where traditional gradient-based methods are impractical or impossible. Our results show that PGT achieves comparable performance to standard backpropagation methods in scenarios where backpropagation cannot be applied, underscoring its value as a training method uniquely suited for such cases.

We estimated that training our reservoir transformer using PGT took approximately 2.8x more training epochs than a comparable traditional transformer. However, the energy efficiency gains achievable with physical reservoirs present a compelling case for adoption. On the higher end of estimates, substituting physical reservoirs into a transformer requires only a 7.5\% improvement in overall energy efficiency to break even after the first year of model deployment. On the lower end, energy reductions of 13-35\% would be necessary. 

Despite its promise, several limitations remain. One notable challenge is the inability to batch training samples, which leads to significantly longer training times per epoch compared to GPU-based systems. Addressing this limitation will require advancements at the reservoir hardware level, as it is not inherently a constraint of PGT itself. Furthermore, integrating physical reservoir transformers into existing AI ecosystems will necessitate improvements in hardware-software co-design to enhance scalability and reduce latency.  The other major hurdle for PGT is that it is currently not able to match the best training losses achieved through current state of the art backpropagation based methods.  Further work needs to be done to address this limitation and refine the method.  Even with these limitation, PGT offers a pathway to significant energy cost savings, potentially reducing AI/ML operational expenses by hundreds of millions of dollars.

On the physics side, future research should focus on optimizing reservoir designs to support parallel processing and exploring hybrid architectures that combine PGT-trained physical reservoirs with GPU-based systems. Such innovations could significantly accelerate training times while preserving the energy efficiency advantages of physical reservoirs.  On the computer science side, further research on PGT should look to refine the training methodology to achieve minimum losses similar to that achieved with backpropagation.

In conclusion, the introduction of Perturbative Gradient Training marks a significant milestone in the field of physical reservoir computing. Much like the impact of stochastic gradient descent (SGD) on the development of modern AI, we hope that PGT will inspire a wave of innovation and research. By bridging the gap between physical reservoirs and state-of-the-art AI systems, this paradigm has the potential to drive sustainable and transformative advancements in artificial intelligence.

\section{Acknowledgments}
Authors acknowledge the support from the National Science Foundation of the United States by Grant No. ECCS-2138236.
\bibliographystyle{IEEEtran}
\bibliography{Re-Ref}

\end{document}